\documentclass[10pt,twocolumn,letterpaper]{article}

\usepackage{ijcb}
\usepackage{times}
\usepackage{epsfig}
\usepackage{graphicx}
\usepackage{amsmath}
\usepackage{amssymb}
\usepackage{multirow} 

\ijcbfinalcopy 

\ifijcbfinal\fi
\begin{document}

\title{Face Super-Resolution with Progressive Embedding of Multi-scale Face Priors}

\author{Chenggong Zhang,  Zhilei Liu\thanks{Corresponding author}\\
College of Intelligence and Computing, Tianjin University\\
Tianjin,  China\\
{\tt\small \{zhangchenggong, zhileiliu\}@tju.edu.cn}
}

\maketitle
\thispagestyle{empty}

\begin{abstract}
   The face super-resolution (FSR) task is to reconstruct high-resolution face images from low-resolution inputs. Recent works have achieved success on this task by utilizing facial priors such as facial landmarks. Most existing methods pay more attention to global shape and structure information, but less to local texture information, which makes them cannot recover local details well. In this paper, we propose a novel recurrent convolutional network based framework for face super-resolution, which progressively introduces both global shape and local texture information. We take full advantage of the intermediate outputs of the recurrent network, and landmarks information and facial action units (AUs) information are extracted in the output of the first and second steps respectively, rather than low-resolution input. Moreover, we introduced AU classification results as a novel quantitative metric for facial details restoration. Extensive experiments show that our proposed method significantly outperforms state-of-the-art FSR methods in terms of image quality and facial details restoration.
\end{abstract}

\section{Introduction}

Recent years have witnessed the rapid progress of technologies for biometrics, especially facial analysis techniques including face recognition and intent recognition from facial expressions~\cite{ren2019research}. Nevertheless, most existing techniques would degrade substantially when given very low-resolution face images. Face super-resolution (FSR), also known as face hallucination, aims to estimate the high-resolution (HR) face images from its degraded low-resolution (LR) observation and restore details. Therefore, FSR can be used as an important means of image preprocessing and greatly benefit face-related tasks which desire high-frequency face details.

As a domain-specific task of general single image super-resolution (SISR), FSR is an inherently ill-posed problem since there are always many possible HR counterparts for every LR image. To reduce the solution space of the problem and promote the performance of FSR, many researchers introduce specific prior information in face images into the FSR problem. A dense correspondence field is employed in~\cite{zhu2016deep} to help restore accurate facial details. Facial landmarks and parsing maps are introduced in~\cite{chen2018fsrnet} to improve recovery performance. A deep iterative collaboration network~\cite{ma2020deep} optimizes face recovery and landmark estimation alternatively. Facial attributes, such as age, gender, and others, are also usually exploited in some face hallucination methods~\cite{lu2018attribute, yu2018super, xin2020facial}. However, most of the existing methods do not take full advantage of different scale prior information and explore only single scale prior information, \ie either global level or local level.

Different faces have distinct distributions in their shape and texture. For face super-resolution, the texture and shape information are both crucial. Many previous face SR methods pay more attention to global shape and structure information, but less to local texture information. Because much prior information such as landmarks and face parsing maps represents the global facial shape. Facial action units (AUs) refer to a set of basic facial muscle actions at certain facial locations defined by the Facial Action Coding System (FACS)~\cite{rosenberg2020face}. The activation or intensity of AUs represents a kind of local semantic and texture information of the face. Some works~\cite{pumarola2018ganimation, liu2020region}, which used AU information for face editing, proved that AU information can affect the local texture of the face. 

To alleviate these above problems, in this paper, we propose a novel recurrent convolutional network based face super-resolution method, which introduces multi-scale facial prior information at different steps progressively. Firstly, we use a branch to estimate landmarks for the output of the first step and input them to the next step to promote facial shape restoration. Then, another branch is used to detect AU information from the output of the second step. In particular, we generate AU attention maps based on the estimated landmarks to better boost the local texture recovery in the following steps. The main contributions of this paper can be summarized as follows:
\begin{itemize}

\item We present a novel FSR framework based on the recurrent network that takes full advantage of multi-scale facial prior information (\ie landmarks and AUs) to generate realistic HR face images.

\item Global shape information and local texture information are embedded into a recurrent network progressively. And we introduced AU classification results as a novel quantitative metric for facial details restoration of FSR.

\item Extensive qualitative and quantitative experiments demonstrate that, compared with similar state-of-the-art methods, our proposed framework achieves superior results in terms of both image quality and facial details restoration.
\end{itemize}

\vspace{-2mm}
\section{Related Work}

\subsection{Single Image Super-Resolution}
FSR is a special case of single image super-resolution (SISR). Recently, due to the excellent learning ability, deep convolutional neural networks have demonstrated high superiority on SISR tasks. Dong \etal~\cite{dong2014learning} firstly presented the SRCNN for SISR and achieved promising performance against traditional methods. Inspired by this pioneering work, many deep network based SISR methods have been proposed. Kim \etal~\cite{kim2016accurate} designed the VDSR network with more convolutional layers based on residual learning~\cite{he2016deep}. Ledig \etal~\cite{ledig2017photo} proposed SRGAN for generating photo-realistic images based on generative adversarial network (GAN)~\cite{goodfellow2014generative}. Shi and Liu~\cite{shi2021dense} proposed DPA-Net for infant fingerprint super-resolution and enhancement. Some attention-based methods~\cite{zhang2018image, dai2019second, liu2020residual} are also proposed to further improve the SR performance. However, most of the above methods have a deep network and hold a lot of parameters, which may suffer from overfitting. To gain better generalization capability without introducing overwhelming parameters, the recurrent structure has also been employed for SISR. Kim \etal~\cite{kim2016deeply} firstly introduced recursive learning in DRCN for parameter sharing. Later, Tai \etal designed a recursive block with enhanced residual units in DRRN~\cite{tai2017image} and memory blocks with the recursive unit and gate unit in MemNet~\cite{tai2017memnet}. Han \etal~\cite{han2018image} presented DSRN considering a dual-state design to exploit features from both LR and HR states for final predictions. Li \etal~\cite{li2019feedback} developed a novel feedback block consisting of  up- and down-sampling layers with dense skip connections in SRFBN. These SISR methods are designed for general images and most of them only handle up to 4× super-resolution. They fail to restore the details well for face images, especially when the scaling factor is large (\eg 8 ×).

\subsection{Face Super-Resolution}
Since the concept of face hallucination was first proposed by Baker and Kanade~\cite{baker2000hallucinating}, many methods were proposed to improve the FSR performance, especially with the development of deep learning. Yu \etal~\cite{yu2016ultra} proposed a GAN-based network URDGN to super-resolve very low-resolution face images. Huang \etal~\cite{huang2017wavelet} presented a wavelet-based method to transform the FSR problem to wavelet coefficients prediction task. Cao \etal~\cite{cao2017attention} proposed Attention-FH using reinforcement learning to discover attended patches and then enhance the facial part sequentially. Dou \etal~\cite{dou2020pca} introduced the incremental orthogonal projection discrimination in the principal component analysis subspace to enhance the FSR task. Lu \etal~\cite{lu2021face} designed a SISN to reconstruct photorealistic high-resolution facial images by fusing the features from two paths.

Compared with general images, the face images have unique prior information which could be utilized. Chen \etal~\cite{chen2018fsrnet} introduced geometry priors including landmark heatmaps and parsing maps. Kim \etal~\cite{kim2019progressive} used landmark heatmaps to design an attention loss. Ma \etal~\cite{ma2020deep} introduced a recurrent network to face SR and designed a deep iterative collaboration framework to optimize face recovery and landmark estimation alternatively. In addition, facial attributes, such as age and gender, are also usually employed in some FSR methods~\cite{lu2018attribute, yu2018super, xin2020facial}. However, most of the existing methods explore only single-level prior information and pay more attention to global shape and structure information, but less to local texture information.


\begin{figure*}[htbp]
\begin{center}
 \includegraphics[width=0.80\linewidth]{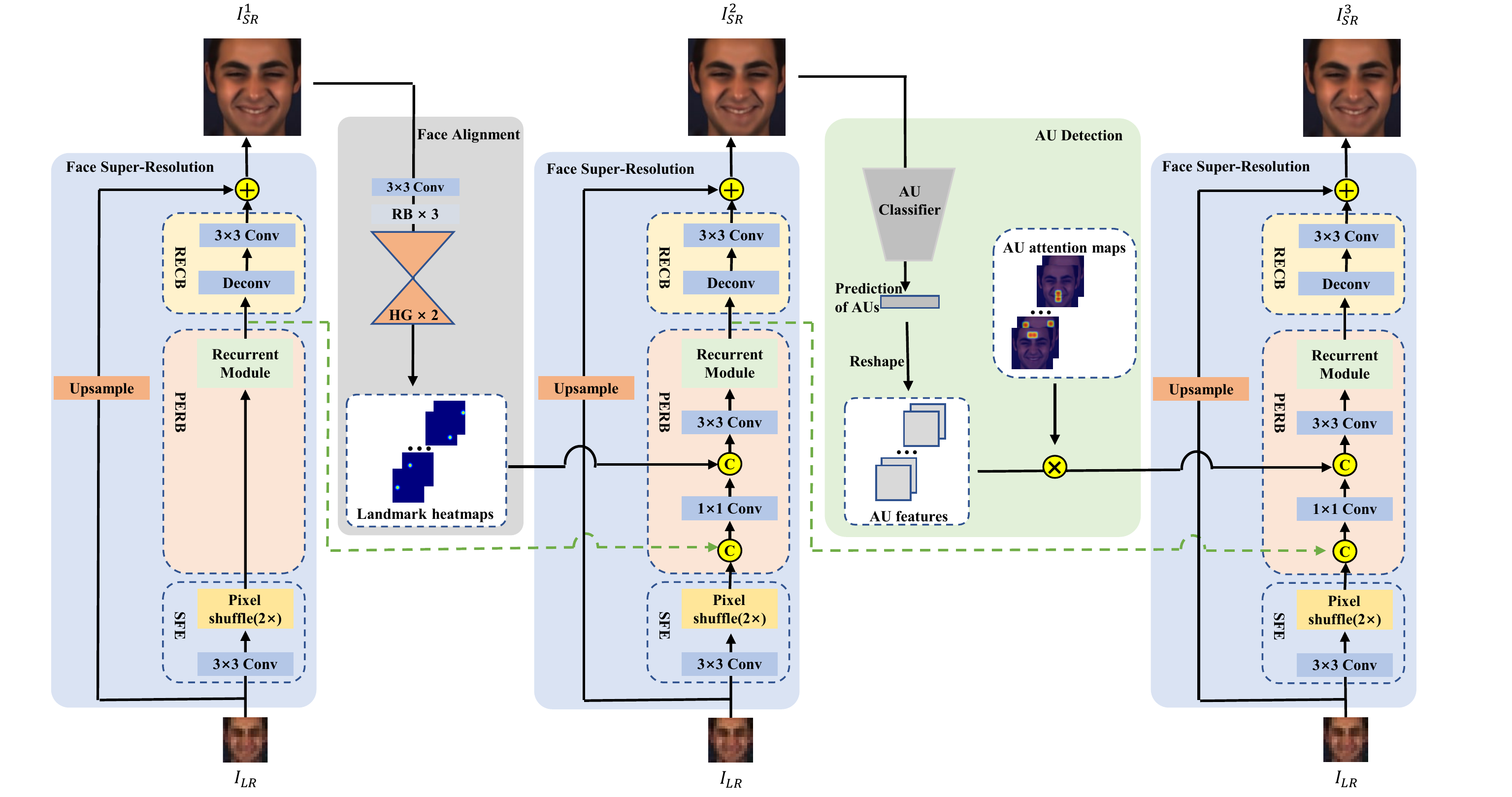}
\end{center}
   \caption{The unfolded architecture of our proposed method, which consists of three branches: face super-resolution branch(blue part), face alignment branch(gray part), and AU detection branch(green part).“C”,"+" and “×” denote concatenation, addition, and multiplication, respectively. The green dotted lines represent feedback connections. For simplicity, we omit the activation function layer in the pipeline.}
\label{fig:model}
\end{figure*}
 
\vspace{-2mm}
\section{Proposed Method}
In this section, we propose a multi-scale prior information embedded recursive network (MPENet) for FSR, which can be unfolded as shown in Figure~\ref{fig:model} and consists of three branches: face super-resolution branch, face alignment branch, and AU detection branch. In addition, to generate photo-realistic face images, we use MPENet as a generator network $G$ and introduce a discriminator network $D$ to build our generative adversarial model MPEGAN.

\vspace{-1mm}
\subsection{Network Architecture}
\textbf{Face Super-Resolution Branch:} As shown in Figure~\ref{fig:model}, our proposed MPENet can be unfolded into 3 iterations. The face super-resolution branch in each iteration contains three parts: a shallow feature extractor (SFE), a prior embedded recurrent block (PERB), and a high resolution reconstruction block (RECB). Given a low-resolution (LR) input $ I_{LR} $, we use a 3×3 convolutional layer and a pixel shuffle layer to extract shallow feature $ F^t_{sf} $ at $ t$-th iteration as:
\begin{equation}
\label{eq_sf}
 F^t_{sf} = H_{SFE}(I_{LR}),
\end{equation}
where $H_{SFE}(\cdot)$ denotes the operations of the SFE. Then, we use a 1×1 and a 3×3 convolutional layer to fuse shallow feature $ F^t_{sf} $, the feedback feature $ F^{t-1}_{fb} $ from previous iteration and prior information $ F_{prior} $, and use a recurrent module to generate a high-level representations $ F^{t}_{fb} $:
\begin{equation}
\label{eq_hiden}
 F^t_{fb} = H_{PERB}(F^t_{sf}, F^{t-1}_{fb}, F_{prior}),
\end{equation}
where $H_{PERB}(\cdot)$ denotes the operations of the PERB. Specifically, the architecture of the recurrent module follows the feedback block in~\cite{li2019feedback} and we remove its first convolutional layer. We set the number of groups to 6 and the number of feature channels to 48. Note that there is no prior information and hidden state at the first iteration in our model. Then, the outputs of the PERB are used as the input to the RECB to generate an SR residual image of the HR face as :
\begin{equation}
\label{eq_res}
 I^t_{Res} = H_{RECB}(F^{t}_{fb}),
\end{equation}
where $H_{RECB}(\cdot)$ denotes the operations of the RECB. RECB consists of a deconvolutional layer and a 3×3 convolutional layer. Finally, the restored SR image at the $t$-th iteration can be described as:
\begin{equation}
\label{eq_sr}
 I^t_{SR} = I^t_{Res} + H_{UP}(I_{LR}),
\end{equation}
where $H_{UP}(\cdot)$ denotes a bilinear upsampling operation. Therefore, we will get totally 3 SR images ($I^1_{SR}$, $I^2_{SR}$, $I^3_{SR}$) for every LR image $I_{LR}$.

\textbf{Face Alignment Branch:} 
Landmarks could represent facial shapes, which provide the locations of facial key points. Because it is difficult to estimate accurate prior information from LR images and recurrent network will generate multiple intermediate SR results, we choose to estimate landmarks from the output $I^1_{SR}$ of the first step. For the face alignment network, we build it with a 3×3 convolutional layer followed by three Residual Blocks (RB)~\cite{he2016deep} and two hourglass (HG) modules~\cite{newell2016stacked}. We impose heatmap loss function $L_{Align}$ for landmark estimation, which is defined as:
\begin{equation}
\label{eq_loss_align}
L_{align} = \mathbb{E}\big [||H_{align}(I^1_{SR}) - L_{gt}||^2_2\big],
\end{equation}
where $H_{align}(\cdot)$ denotes the operation of alignment network and $L_{gt}$ denotes the ground-truth heatmap.


\textbf{AU Detection Branch:} 
Facial action units (AUs) refer to a set of basic facial muscle actions at certain facial locations and represent local texture information of the face. Different facial AUs have different scales. Large-scale AUs can be better restored with the help of landmark information. AU classifier is designed to promote the restoration of small-scale AUs information through AU loss $L_{au}$ as much as possible. Here, we use ResNet-18~\cite{he2016deep} as AU classifier to detect AUs from the output $I^2_{SR}$ of the second step. The weighted multi-label cross-entropy loss is used to calculate the loss between the predicted AUs and the ground-truth AU labels as:
\begin{equation}
\label{eq_loss_cross}
L_{cross} = -{\sum_{i=1}^{n_{au}} w_i[\hat{y}_ilog\hat{y}_i + (1 - {y}_i)log(1 - \hat{y}_i)]},
\end{equation}
where $y_i$ denotes the ground-truth label of the $i$-th AU, $\hat{y}_i$ represents the corresponding predicted occurrence probability and $n_{au}$ is the number of AUs. In many cases, some AUs occur rarely in samples~\cite{shao2021jaa}, for which the network often predicts the AU strongly biased towards non-occurrence. To mitigate this issue, a weighted multi-label Dice coefficient loss~\cite{milletari2016v} is introduced:
\begin{equation}
\label{eq_loss_dice}
L_{dice} = \sum_{i=1}^{n_{au}} w_i(1 - \frac{2y_i\hat{y}_i + \epsilon}{y^2_i + \hat{y}^2_i + \epsilon}),
\end{equation}
where $\epsilon$ is the smooth term. The weight $w_i = \frac{(1/r_i)n_{au}}{\sum_{i=1}^{n_{au}}1/r_i}$ introduced in Eq. (\ref{eq_loss_cross}) and Eq. (\ref{eq_loss_dice}) is to alleviate the data imbalance problem~\cite{martinez2017automatic}, where $r_i$ is the occurrence rate of the $i$-th AU in the training set. Thus, the AU loss used in our method is:
\begin{equation}
\label{eq_loss_au}
L_{au} = L_{cross} + L_{dice},
\end{equation}

To take full advantage of AUs information, the predicted AUs are embedded into PERB  at the third iteration to correct unreasonable facial details. To facilitate the embedding operation, we reshape the predicted AU values into a certain number of feature maps, in which each channel corresponds to an AU. Furthermore, for each AU to affect only its corresponding region, we utilize the landmarks estimated in the previous step to generate AU attention maps following the rules in~\cite{shao2021jaa} and then multiply the AU feature maps by them.

\vspace{-1mm}
\subsection{Loss Functions}
In addition to landmark heatmap loss and AU loss mentioned above, reconstruction loss, perceptual loss, and adversarial loss are also considered for training our model.

\textbf{Reconstruction Loss:} To achieve better optimization, we impose pixel-wise loss to each output of $ N $ iterations, which can be formulated as:
\begin{equation}
\label{eq_loss_rec}
L_{rec} = \mathbb{E}\bigg[\frac{1}{N}\sum_{n=1}^{N}||I_{HR}-I^n_{SR}||^2_2 \bigg],
\end{equation}
where $I_{HR}$ denotes the ground-truth HR images.

\textbf{Perceptual Loss:} 
To obtain HR images with better perceptual quality, a perceptual loss~\cite{johnson2016perceptual} is also imposed. We adopt the pre-trained VGG-Face network~\cite{parkhi2015deep} to extract high-level features (\ie, features from the ‘relu5\_3’ layer) from SR and HR images. The perceptual loss is defined as:

\begin{equation}
\label{eq_loss_per}
L_{per} = \mathbb{E}\big[||\phi(I_{HR})-\phi(I^3_{SR})||^2_2 \big],
\end{equation}
where $\phi$ denotes the operation of mapping the images to the feature space via the pre-trained VGG-Face model.

\textbf{Adversarial Loss:} 
Inspired by the recent success of GAN in super-resolution~\cite{ledig2017photo, chen2018fsrnet, ma2020deep}, we also introduce the adversarial loss~\cite{ledig2017photo} to improve the reality of generated face images. The loss functions of the generator $G$ and discriminator $D$ are formulated as:

\begin{equation}
\label{eq_loss_advg}
L_{adv\_G} = - \mathbb{E}[log(D(G(I_{LR})))],
\end{equation}

\begin{equation}
\label{eq_loss_advd}
L_{adv\_D} = -\mathbb{E}[log(D(I_{HR}))] - \mathbb{E}[log(1 - D(G(I_{LR})))].
\end{equation}

\textbf{Overall Loss:} 
Thus, the final overall loss term of our generator is shown as:

\begin{equation}
\begin{aligned}
\label{eq_loss_all}
L_{overall} = L_{rec} + \lambda_{align}L_{align} + \lambda_{au}L_{au} \\ + \lambda_{per}L_{per} + \lambda_{adv}L_{adv\_G},
\end{aligned}
\end{equation}
where $\lambda_{align}, \lambda_{au}, \lambda_{per}, \lambda_{adv}$ are trade-off parameters. For the training of our PSNR-oriented model MPENet, we set $\lambda_{per} = \lambda_{adv} = 0$. Then we utilize complete losses to train the perceptual-pleasing model MPEGAN.

\begin{figure*}
\begin{center}
 \includegraphics[width=0.80\linewidth]{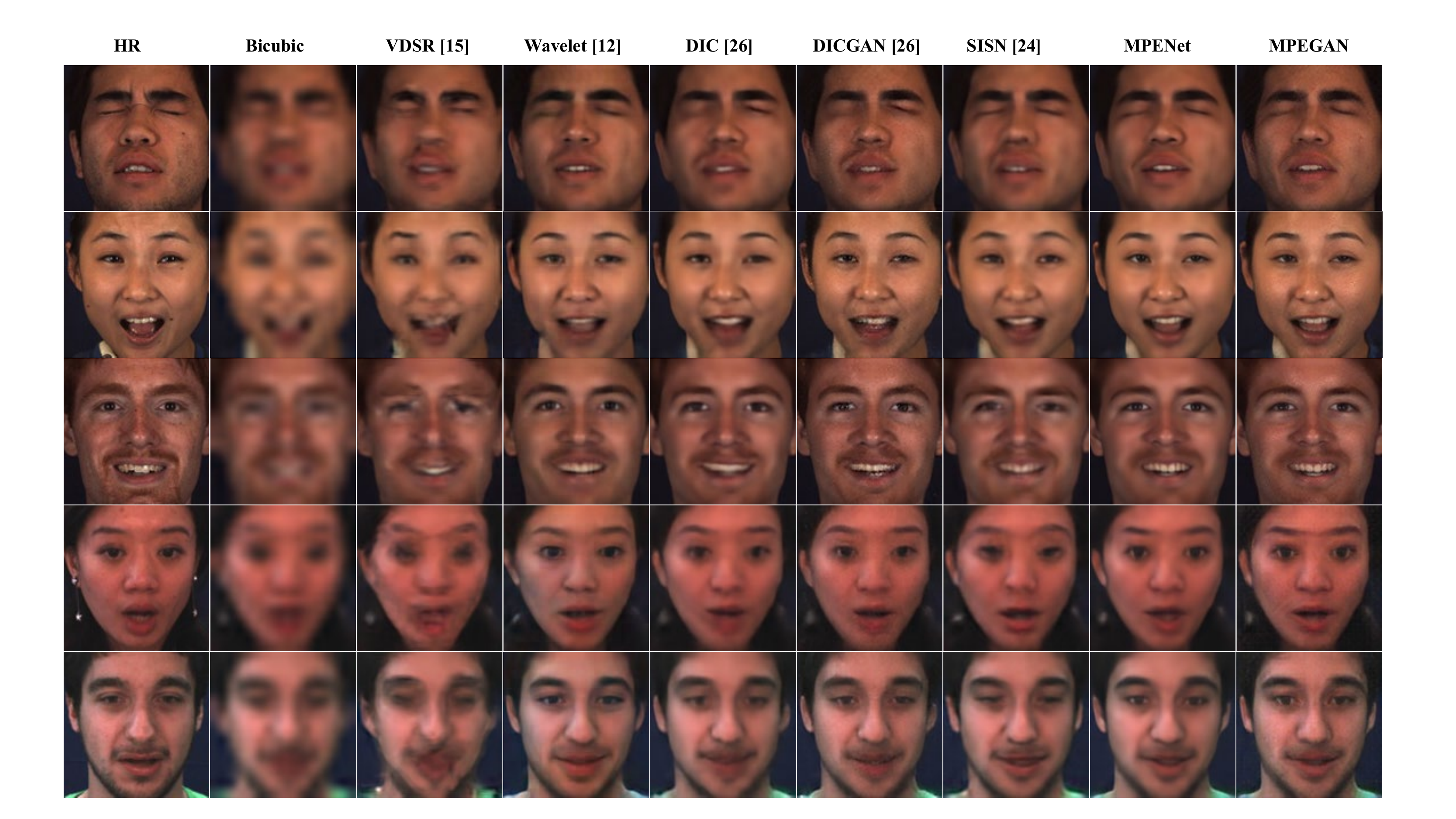}
\end{center}
   \caption{Qualitative comparison with state-of-the-art methods in 8× SR. Top three examples are of BP4D and others are of DISFA. Best viewed on screen.}
\label{fig:Qualitative}
\end{figure*}

\vspace{-2mm}
\section{Experiments}

\subsection{Datasets and Metrics}
\textbf{Datasets:} We conduct extensive experiments on two face datasets, \ie BP4D~\cite{zhang2014bp4d} and DISFA~\cite{mavadati2013disfa}, in which AU labels are provided. 
There are around 140,000 frames with labels of 12 AUs in the BP4D dataset. DISFA dataset consists of 27 subjects, each of which has 4,845 frames, with intensities annotated of 8 AUs in the range from 0 to 5. The frames with intensities equal to or greater than 2 are considered positive, while others are treated as negative.  For BP4D, we follow the settings of~\cite{li2017eac, shao2021jaa} and use subject exclusive 3-fold cross-validation. For DISFA, we divide the dataset into a training set (18 subjects) and a test set (9 subjects).
For both datasets, we use Dlib toolkit~\cite{king2009dlib} to detect 68 landmarks as the ground-truth.

\textbf{Evaluation Metrics:} We evaluate SR results with the widely used quantitative
metrics including Peak Signal to Noise Ratio (PSNR) and structural similarity (SSIM)~\cite{wang2004image}. They are computed on the luminance (Y) channel of the YCbCr space. We also use the classification results of AUs as a metric to evaluate the accuracy of facial details restoration. We use OpenFace~\cite{baltrusaitis2013constrained, baltrusaitis2018openface, zadeh2017convolutional} to detect the AUs and use F1-score(\%) and accuracy (Acc, \%) to evaluate AU classification results. We calculate the average results of all AUs. In the display of BP4D, all the quantitative results are the average over the trifold dataset. For the sake of simplicity, \% in all the results is omitted in the following sections.

\vspace{-1mm}
\subsection{Implementation Details}
The similarity transformation, including rotation, uniform scaling, and translation, is imposed on each image to obtain $144\times144$ face image. 
This transformation is shape-preserving and brings no change to the texture details. To improve the network robustness, we augment training images by randomly cropping  $128\times128$ patch from each transformed image, and horizontally flipping. For testing, images are only center-cropped to $128\times128$. Then we obtain $16\times16$ LR inputs by bicubic downsampling. The model structure is shown in Figure~\ref{fig:model}. Unless otherwise specified, the number of layers is 1. We train the MPENet with the weights $\lambda_{align}=0.1$ and $\lambda_{au}=0.01 $. For MFEGAN training, we use the parameters of pre-trained MPENet as initialization and set the coefficients as: $\lambda_{align}=\lambda_{per}=0.1, \lambda_{au}=0.01$ and $\lambda_{adv\_G}=0.001$. We employ ADAM optimizer~\cite{kingma2014adam} to train both models end-to-end with $\beta_{1} = 0.9$, $\beta_{2}$ = 0.999 and $\epsilon=10^{-8}$, and all branches are jointly trained for 2 epochs. The batch size is set to 4. The learning rate is $1\times10^{-4}$. Our experiments are implemented on Pytorch framework~\cite{paszke2017automatic} with an NVIDIA Titan V GPU.


\vspace{-1mm}
\subsection{Comparison with the state-of-the-arts}
We compare our proposed method with state-of-the-art SR methods, including generic SR methods like SRCNN~\cite{dong2014learning}, VDSR~\cite{kim2016accurate}; and face SR methods like WaveletSRNet~\cite{huang2017wavelet}, DIC~\cite{ma2020deep}, and SISN~\cite{lu2021face}. Bicubic interpolation is also introduced as a baseline. For a fair comparison, all models are trained on the same dataset using the published codes of the above methods. Table~\ref{tab:PSNR_SSIM} presents the quantitative results on the test set of BP4D and DISFA. As we can see, our MPENet outperforms other methods on PSNR and SSIM on both datasets. What is noteworthy is that our model achieves better performance than DIC which only utilizes landmark information. In addition, our MPEGAN maintains high PSNR and SSIM than DICGAN while improving the authenticity and perceptual quality of SR images. These indicate that introducing both global shape and local texture information can promote FSR performance.
\begin{table}
\footnotesize
\begin{center}
\scalebox{0.90}{
\begin{tabular}{|c|c|c|c|c|}
\hline
 
{\multirow{2}*{Method}}&\multicolumn{2}{c|}{BP4D}&\multicolumn{2}{c|}{DISFA}\\
\cline{2-5}
&PSNR&SSIM&PSNR&SSIM\\
\hline\hline

{Bicubic}& 28.67 & 0.7724 & 28.88 & 0.8031 \\
\hline
{SRCNN~\cite{dong2014learning}}& 30.12 & 0.8070 & 29.71 & 0.8193\\
\hline
{VDSR~\cite{kim2016accurate}}& 30.27 & 0.8236 & 29.47 & 0.8192\\
\hline
{Wavelet~\cite{huang2017wavelet}}& 31.25 & 0.8600 & 30.70 & 0.8773\\
\hline
{DIC~\cite{ma2020deep}}& 31.74 & 0.8648 & 31.40 & 0.8714 \\
\hline
{DICGAN~\cite{ma2020deep}}& 31.08 & 0.8392 & 31.09 & 0.8548\\
\hline
{SISN~\cite{lu2021face}}& 32.02 & 0.8683 & 31.78 & 0.8723\\
\hline
{MPENet}&\bf32.22 & \bf0.8765 & \bf32.02 & \bf0.8841\\
\hline
{MPEGAN}& 31.43 & 0.8364 & 31.46 & 0.8566\\
\hline

\end{tabular}

}
\end{center}
\caption{Comparison (PSNR and SSIM) of various methods in 8× SR. The best results are \textbf{highlighted}.}
\label{tab:PSNR_SSIM}
\end{table}

Figure~\ref{fig:Qualitative} provides some SR results of various methods on two datasets, which can provide an intuitive observation. It can be observed that our MPENet reconstructs fine facial details while others fail to generate satisfactory results. The important facial structures and details also look visually similar to the HR images. For example, the eye region of the results of the first row and the mouth region of the results of the last row.  In contrast, we can also see from Table~\ref{tab:PSNR_SSIM} and Figure~\ref{fig:Qualitative} that the SISN obtains the second-best PSNR and SSIM values, but it tends to generate smooth results with fewer details. Furthermore, our MPEGAN can generate perceptually better SR results.   

Moreover, because some details of the differences are not easy to observe, we conduct facial AU classification results as a metric to measure the details restoration of SR images. The F1-score and accuracy are reported to evaluate the performance. To better reflect the effect of introducing multi-scale semantic information on detail recovery, we omit GAN-based methods when calculating them. In Table~\ref{tab:au}, it can be seen that our MPENet achieves the best results on both datasets. It is noteworthy that Wavelet gets comparable results with our MPENet, which designed a texture loss to train their model. We think that the texture loss has a positive effect on the restoration of face texture, but their method performs lower PSNR and SSIM as can be seen in Table~\ref{tab:PSNR_SSIM}. Differently, our method can recover facial details while preserving pixel-wise accuracy due to the embedding of multi-scale prior information.

\begin{table}
\begin{center}
\footnotesize
\scalebox{0.90}{
\begin{tabular}{|c|c|c|c|c|}
\hline
 
{\multirow{2}*{Method}}&\multicolumn{2}{c|}{BP4D}&\multicolumn{2}{c|}{DISFA}\\
\cline{2-5}
&F1-score&Acc&F1-score&Acc\\
\hline\hline

{Bicubic}& 52.98 & 69.39 & 34.65 & 71.73 \\
\hline
{SRCNN~\cite{dong2014learning}}& 55.74 & 69.79 & 39.44 & 73.14\\
\hline
{VDSR~\cite{kim2016accurate}}& 56.31 & 71.47 & 40.14 & 71.99\\
\hline
{Wavelet~\cite{huang2017wavelet}}& 58.83 & 73.90 & 44.18 & 73.11\\
\hline
{DIC~\cite{ma2020deep}}& 57.19 & 73.51 & 45.59 & 72.98 \\
\hline

{SISN~\cite{lu2021face}}& 58.42 & 73.46 & 45.27 & 72.95\\
\hline
{MPENet}&\bf58.84 & \bf74.27 & \bf46.38 & \bf73.29\\
\hline

\end{tabular}
}
\end{center}
\caption{Comparison (F1-score and Accuracy) of various methods in 8× SR. The best results are \textbf{highlighted}.}
\label{tab:au}
\end{table}



\vspace{-1mm}
\subsection{Model Analysis}
Firstly, we investigated the effects of introducing landmarks and AUs on the SR output of different iterative steps. The quantitative results on BP4D and DISFA are tabulated in Table~\ref{tab:step}. We can see that the performance gets better progressively as the number of steps increases. However, when $t$ increases to 3, PSNR and SSIM tend to be stable. Therefore, considering computational efficiency and SR performance, the results of the third step may be an appropriate choice for our method.

\begin{figure}[t]
\begin{center}
   \includegraphics[width=0.79\linewidth]{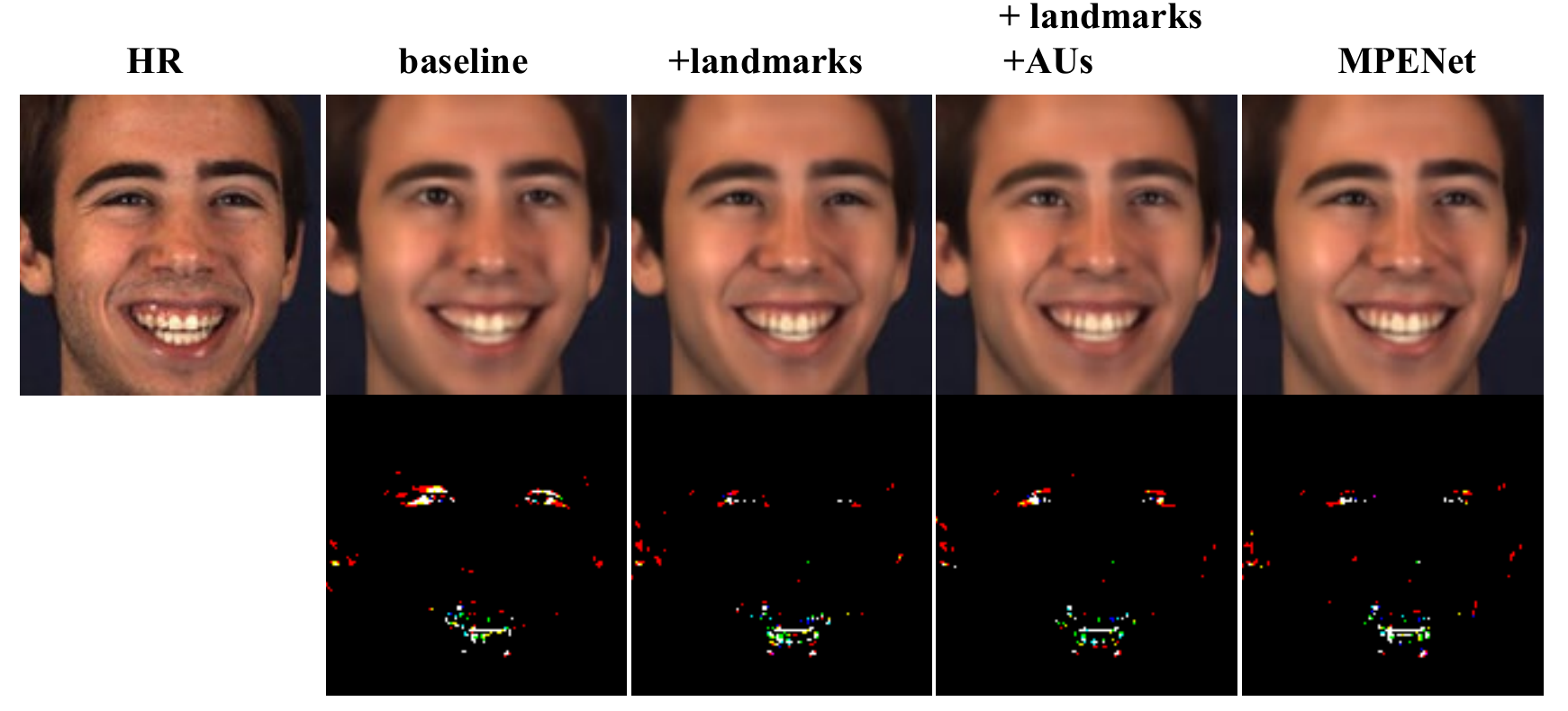}
\end{center}
   \caption{Ablation study on effects of landmarks, AUs, and AU
attention maps. The second line shows the normalized difference map between each result and HR. Best viewed on screen and zoom in to see the details.}
\label{fig:ablation}
\end{figure}

In addition, to explore the effectiveness of each component of our method, an ablation study is further implemented on the BP4D dataset. Our base model is denoted as the baseline that removes the face alignment branch and AU detection branch. When a branch is removed, its corresponding loss function is also removed. All variants in our ablation study are trained
for the same number of stages. As shown in Table~\ref{tab:ablation}, PSNR and SSIM have been improved after adding landmarks, AUs, and AU attention maps progressively. Overall, F1-score and accuracy also tend to improve, but we see a slight negative impact on the F1-score when AU information is added without attention maps. The reason is that directly concatenating reshaped AU features with feature maps will affect other facial regions except the corresponding of each AU.  Finally, when we introduced AU attention maps, all the metrics achieved the best results. Moreover, we also give the qualitative results of each variant in Figure~\ref{fig:ablation}. It can be seen that with the addition of each component, the details become more accurate, such as the eye area. These experiments indicate the effectiveness of each component of our proposed method.

%
\begin{table}
\begin{center}
\footnotesize
\scalebox{0.90}{
\begin{tabular}{|cc|c|c|c|c|}
\hline
\multicolumn{2}{|c|}{Dataset}                        & PSNR           & SSIM            & F1-score       & Acc            \\ \hline \hline
\multicolumn{1}{|c|}{\multirow{3}{*}{BP4D}}  & step1 & 31.70          & 0.8606          & 57.14          & 72.98          \\ \cline{2-6} 
\multicolumn{1}{|c|}{}                       & step2 & 32.21          & 0.8763          & 58.74          & 74.24          \\ \cline{2-6} 
\multicolumn{1}{|c|}{}                       & step3 & \textbf{32.22} & \textbf{0.8765} & \textbf{58.84} & \textbf{74.27} \\ \hline \hline
\multicolumn{1}{|c|}{\multirow{3}{*}{DISFA}} & step1 & 30.84          & 0.8513          & 45.53          & 72.24          \\ \cline{2-6} 
\multicolumn{1}{|c|}{}                       & step2 & 32.02          & 0.8841          & 46.02          & 72.83          \\ \cline{2-6} 
\multicolumn{1}{|c|}{}                       & step3 & \textbf{32.02} & \textbf{0.8841} & \textbf{46.38} & \textbf{73.29} \\ \hline
\end{tabular}
}
\end{center}
\caption{Quantitative evaluations of different iterative steps on both datasets. The best results are \textbf{highlighted}.}
\label{tab:step}
\end{table}

\begin{table}
\begin{center}
\footnotesize
\scalebox{0.90}{
\begin{tabular}{|c|c|c|c|c|}
\hline
 
{Model}& PSNR & SSIM & F1-score & Acc \\

\hline\hline

{baseline}& 31.87 & 0.8662 & 57.79 & 72.33 \\
\hline
{+landmarks}& 32.12 & 0.8747 & 58.72 & 74.15\\
\hline
+landmarks&{\multirow{2}*{32.15}} &{\multirow{2}*{0.8755}} &{\multirow{2}*{58.64}} &{\multirow{2}*{74.21}}\\
+AUs&&&&\\
\hline

+landmarks&{\multirow{3}*{\bf32.22}} &{\multirow{3}*{\bf0.8765}} &{\multirow{3}*{\bf58.84}} &{\multirow{3}*{\bf74.27}}\\
+AUs&&&&\\
+attention maps&&&&\\
\hline

\end{tabular}
}
\end{center}
\caption{Ablation study on effects of landmarks, AUs, and AU attention maps on BP4D dataset. The last row is our MPENet. The best results are \textbf{highlighted}.}
\label{tab:ablation}
\end{table}

\vspace{-2mm}
\section{Conclusion}
In this paper, we propose a novel recurrent convolutional network based framework for face super-resolution, which introduces both global shape and local texture information progressively to promote the performance. Landmarks information and facial action units (AUs) information are extracted in the output of the first and second steps respectively, rather than LR input. The experimental results on the BP4D dataset and DISFA dataset demonstrate that landmarks and AUs are beneficial to recovering facial structure and local texture details, and our presented method is superior to the state-of-the-art in terms of both AUs classification and image quality. In future work, we will explore the relationship between multi-scale information to promote their better integration.
\section*{Acknowledgment}
\noindent
This work is supported by the National Natural Science Foundation of China (61503277) and the National Key Research and Development Program (2019YFE0198600).

{\small
\bibliographystyle{ieee}
\bibliography{egbib}
}

\end{document}